\pdfoutput=1

\documentclass[11pt]{article}
\usepackage{acl}

\usepackage{times}
\usepackage{latexsym}
\usepackage[T1]{fontenc}
\usepackage[utf8]{inputenc}
\usepackage{microtype}
\usepackage{inconsolata}
\usepackage{graphicx}
\usepackage{xcolor}
\usepackage{paralist}
\usepackage{booktabs}
\usepackage{subcaption} 
\usepackage{algorithm}
\usepackage{algorithmicx}
\usepackage{algpseudocode}
\usepackage{comment}
\usepackage{hyperref}
\usepackage{amsmath}
\usepackage{tabularx}
\usepackage{stfloats}
\usepackage{enumitem} 
\title{Pre-trained Models Perform the Best \\ When Token Distributions Follow Zipf's Law}

\author{Yanjin He\textsuperscript{\rm 1,\rm 2}, Qingkai Zeng\textsuperscript{\rm 2,\rm 3}\footnotemark[1], Meng Jiang\textsuperscript{\rm 2} \\
        \textsuperscript{\rm 1} School of Mathematical Sciences, Peking University \\
        \textsuperscript{\rm 2} Department of Computer Science and Engineering, University of Notre Dame \\
        \textsuperscript{\rm 3} College of Computer Science, Nankai University \\
        \ yanjinhe@stu.pku.edu.cn, qzengnkcs@gmail.com\\
        mjiang2@nd.edu \\}

\begin{document}
\maketitle

\renewcommand{\thefootnote}{\fnsymbol{footnote}}
\footnotetext[1]{Corresponding author.}
\renewcommand{\thefootnote}{\arabic{footnote}}

\begin{abstract}
Tokenization is a fundamental step in natural language processing (NLP) and other sequence modeling domains, where the choice of vocabulary size significantly impacts model performance. Despite its importance, selecting an optimal vocabulary size remains underexplored, typically relying on heuristics or dataset-specific choices. In this work, we propose a principled method for determining the vocabulary size by analyzing token frequency distributions through Zipf’s law. We show that downstream task performance correlates with how closely token distributions follow power-law behavior, and that aligning with Zipfian scaling improves both model efficiency and effectiveness. Extensive experiments across NLP, genomics, and chemistry demonstrate that models consistently achieve peak performance when the token distribution closely adheres to Zipf’s law, establishing Zipfian alignment as a robust and generalizable criterion for vocabulary size selection. The code and data are available at: \url{https://github.com/yanjinhe/Tokenizer}
\end{abstract}

\section{Introduction}

Tokenization is a fundamental preprocessing step in natural language processing (NLP), where raw text is segmented into smaller units known as tokens \cite{sennrich2016neural}. These tokens can represent words, subwords, or characters, depending on the tokenization strategy \cite{schuster2012japanese}, and they form the basis for subsequent representation learning. The choice of tokenizer and its vocabulary size has a direct impact on model capacity, robustness, and computational efficiency \cite{devlin2019bert}.

Among various strategies, Byte Pair Encoding (BPE) \cite{sennrich2016neural} is the most widely adopted method in modern large language models. Existing large language models typically fix a vocabulary size (e.g., 50K) \cite{achiam2023gpt} in advance, then apply BPE to construct the tokenizer. This fixed-size approach, while convenient, lacks a principled basis and may not be optimal across different tasks, domains, or languages.

In practice, choosing too small a vocabulary may lead to fragmented or overly fine-grained tokens, resulting in longer sequences and degraded semantic representation\cite{provilkov2020bpe}. On the other hand, overly large vocabularies may introduce redundancy, inflate memory usage, and reduce model efficiency \cite{brown2020language}. However, vocabulary size is often treated as a fixed hyperparameter, determined heuristically or based on dataset statistics \cite{kudo2018sentencepiece}.

Several prior works have explored metrics such as fertility (token-per-word ratio), parity (cross-lingual symmetry), and coverage to evaluate tokenizers \cite{liu2020multilingual,wu2016google}. However, these metrics have been shown to correlate poorly with downstream task performance\cite{ali2024tokenizer}, especially when moving beyond NLP to other modalities such as genomics or chemistry. As a result, there remains a need for a more robust criterion to guide vocabulary size selection.

In this work, we propose a principled approach inspired by Zipf’s law, a well-known linguistic phenomenon whereby word frequency is inversely proportional to its rank in natural language corpora \cite{powers1998applications}. We hypothesize that effective tokenizers should induce token frequency distributions that align with Zipfian behavior. To test this hypothesis, we introduce the \textit{Zipf alignment score}, which quantifies how closely a tokenizer's frequency distribution fits a power-law on a log-log plot. We use this score as a proxy metric to guide vocabulary size selection.

Empirically, we demonstrate that token distributions adhering more closely to Zipf’s law correspond to better downstream performance. Our experiments span NLP, genomics, and chemistry tasks, showing that Zipf alignment consistently predicts optimal vocabulary size across modalities. 

To summarize, the main contributions of our paper are as follows:
\begin{itemize}[leftmargin=*, noitemsep, topsep=0pt]
    \item We show that as the vocabulary size increases, the token frequency distribution on a log-log scale becomes increasingly linear, reflecting stronger alignment with Zipf’s law.
    \item We demonstrate that downstream task performance consistently improves and reaches its peak when the token distribution most closely follows Zipfian behavior.
    \item We propose a principled approach for selecting vocabulary size by measuring the degree of Zipf alignment in the token distribution. This method is simple, generalizable across domains, and predictive of optimal performance.
\end{itemize}

\section{Related Work}
\label{sec:related_work}

\paragraph{Tokenization} Tokenization, the process of segmenting raw data into smaller units, is a critical step in NLP and other fields. Classic methods like BPE \cite{sennrich2016neural} and WordPiece \cite{schuster2012japanese} use subword segmentation to balance vocabulary size and out-of-vocabulary handling, while SentencePiece \cite{kudo2018sentencepiece} enables language-independent tokenization. These methods are foundational for modern models like BERT \cite{devlin2019bert} and GPT \cite{radford2019language}, as tokenization directly impacts model efficiency, robustness, and downstream task performance. Beyond text, tokenization has been adapted for genomics (e.g., k-mer tokenization in DNABERT \cite{ji2021dnabert}), chemistry (e.g., SMILES segmentation \cite{schwaller2019molecular}), and even vision and audio, where images are split into patches and audio into spectrograms \cite{dosovitskiy2021image, radford2023robust}, demonstrating its versatility across modalities.

\paragraph{Tokenizer Selection Criteria}
Prior work has explored several heuristics for selecting tokenizers. One common approach is to use compression ratio as a proxy, under the assumption that better compression implies more efficient representations. \citet{goldman2024unpacking} examine this hypothesis and find that compression correlates with performance in some cases, but not consistently. \citet{ali2024tokenizer} further evaluate metrics such as fertility, parity, and compression, showing that these do not reliably predict downstream task performance. These findings suggest that standard metrics often fail to capture what makes a tokenizer effective, highlighting the need for more robust, task-aware criteria.

\paragraph{Zipf's law and Power Law} Power-law distributions were first studied by Pareto in the context of wealth distribution \cite{pareto1964cours}. Zipf later formalized this phenomenon in linguistics, showing that word frequencies in natural language follow a power-law distribution, now known as Zipf's law \cite{zipf2013psycho}. This distribution reveals that a small number of words dominate the text frequency, while most words are uncommon, a pattern that is consistent across languages and corpora \cite{montemurro2001beyond}. Power-law distributions are also prevalent in other domains, including biology, where gene expression levels and protein networks exhibit scaling laws \cite{jeong2001lethality}, and in social networks, where the degree distribution of connections follows power-law behavior \cite{barabasi1999emergence}.

\section{Observing Zipf's Law}
\label{sec:preliminary}
One of the most widely adopted subword tokenization methods is Byte Pair Encoding (BPE) \cite{gage1994new}, which iteratively merges the most frequent adjacent character pairs in a corpus until a predefined vocabulary size is reached. The BPE algorithm is shown in Appendix ~\ref{sec:appendix1}. BPE has been extensively used in state-of-the-art large-scale language models. Given its widespread adoption, BPE shows its importance in NLP research.

\subsection{Vocabulary Size}

Vocabulary size is a critical yet often overlooked factor in designing tokenizers. If a model is trained on an infinitely large dataset that comprehensively represents all knowledge, and if the model has access to unlimited computational resources, then vocabulary size is of minimal concern—one can simply choose a sufficiently large vocabulary. However, in real-world scenarios, training datasets represent only a subset of global knowledge, and computational resources impose practical limitations on training. This makes vocabulary size an essential hyperparameter.

A small vocabulary set may fail to capture the fundamental characters of a dataset, leading to excessive fragmentation of words and loss of semantic information. Conversely, an overly large vocabulary set would introduce redundancy, leading to inefficient token representations that are not optimally compact. This trade-off is especially pronounced when dealing with domain-specific datasets, where suboptimal vocabulary choices can significantly impact model performance.

Despite its importance, vocabulary size is often determined based on heuristics or set arbitrarily large without systematic optimization. Such arbitrary choices may prevent models from capturing the most meaningful token distributions for a given dataset, potentially limiting performance. We argue that optimal vocabulary size should be carefully determined for each dataset, particularly in different modalities such as NLP, genomics, and chemistry. Identifying the appropriate vocabulary size for a given domain is crucial for maximizing information retention and model efficiency.
\subsection{Power Law and Token Rank-Frequency Distributions}

Power law distributions characterize many naturally occurring phenomena, including linguistic structures. A power law describes a relationship where the frequency of an event is inversely proportional to its rank, typically expressed as $ f(x) \propto x^{-k}$, where $x$ is the rank \cite{pareto1964cours}.

The log-log token rank-frequency distribution is based on empirical observations of textual data and is used to analyze the probabilistic structure of word frequencies within a text or corpus. In this representation, both the frequency of tokens (words) and their rank by frequency are plotted on logarithmic scales. If the token frequency follows a perfect power-law distribution, the plot should form a straight line. However, in many real-world datasets, as shown in Section~\ref{sec:experiments} and Figure~\ref{fig:zipf}, the plot often consists of segments with different slopes, indicating the presence of multiple classes of tokens with varying degrees of redundancy. 

In natural language, we typically observe that vocabulary distributions follow a power-law when trained on sufficiently large datasets. This observation motivates us to investigate token distribution patterns, particularly in specific datasets or domains. It leads us to ask: \textit{What is the optimal token distribution for a given domain or dataset? Can we determine the vocabulary size prior to training to obtain such an optimal distribution?}

\subsection{Hypotheses and Vocabulary Size Selection Strategy}
\label{sec:zipf_hypothesis}

Our study begins with the empirical observation that the \textit{token rank-frequency distribution} exhibits a Zipfian pattern. This leads us to propose two hypotheses that guide our vocabulary size selection:

\begin{compactenum}
    \item \textbf{Hypothesis 1:} As vocabulary size increases, the log-log rank-frequency distribution of tokens gradually approaches a straight line, indicating alignment with Zipf's law.
    \item \textbf{Hypothesis 2:} When the token distribution closely matches Zipf’s law, the model achieves superior downstream performance.
\end{compactenum}

In this section, we focus on verifying \textbf{Hypothesis 1} using the BookCorpus dataset. We train BPE-based tokenizers with various vocabulary sizes (ranging from 2K to 50K) and visualize the resulting rank-frequency distributions in log-log space.

\begin{figure*}[t]
    \centering
    \begin{subfigure}[b]{0.32\textwidth}
        \includegraphics[width=\textwidth]{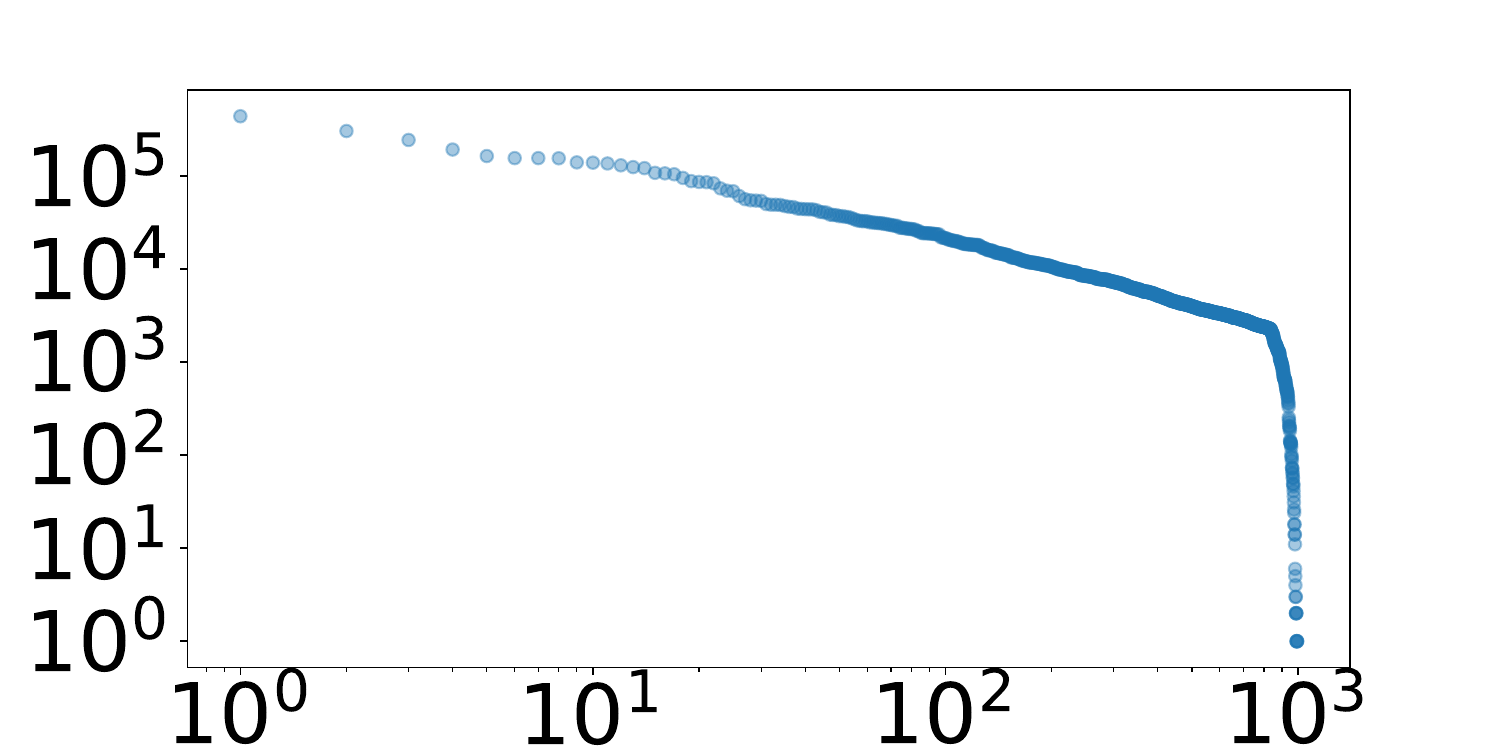}
        \caption{Vocab Size=1000}
        \label{fig:zipf1}
    \end{subfigure}
    \hfill 
    \begin{subfigure}[b]{0.32\textwidth}
        \includegraphics[width=\textwidth]{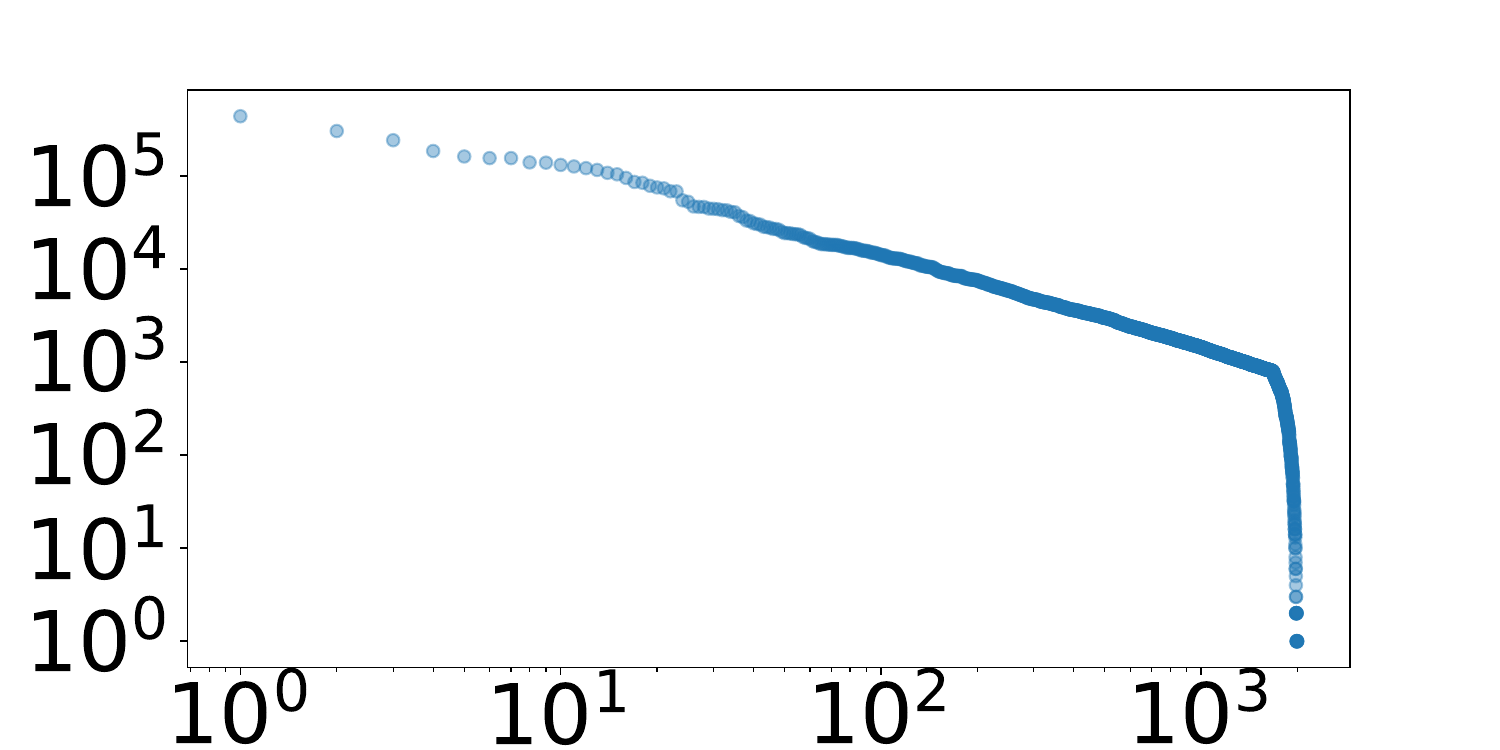}
        \caption{Vocab Size=2000}
        \label{fig:zipf2}
    \end{subfigure}
    \hfill
    \begin{subfigure}[b]{0.32\textwidth}
        \includegraphics[width=\textwidth]{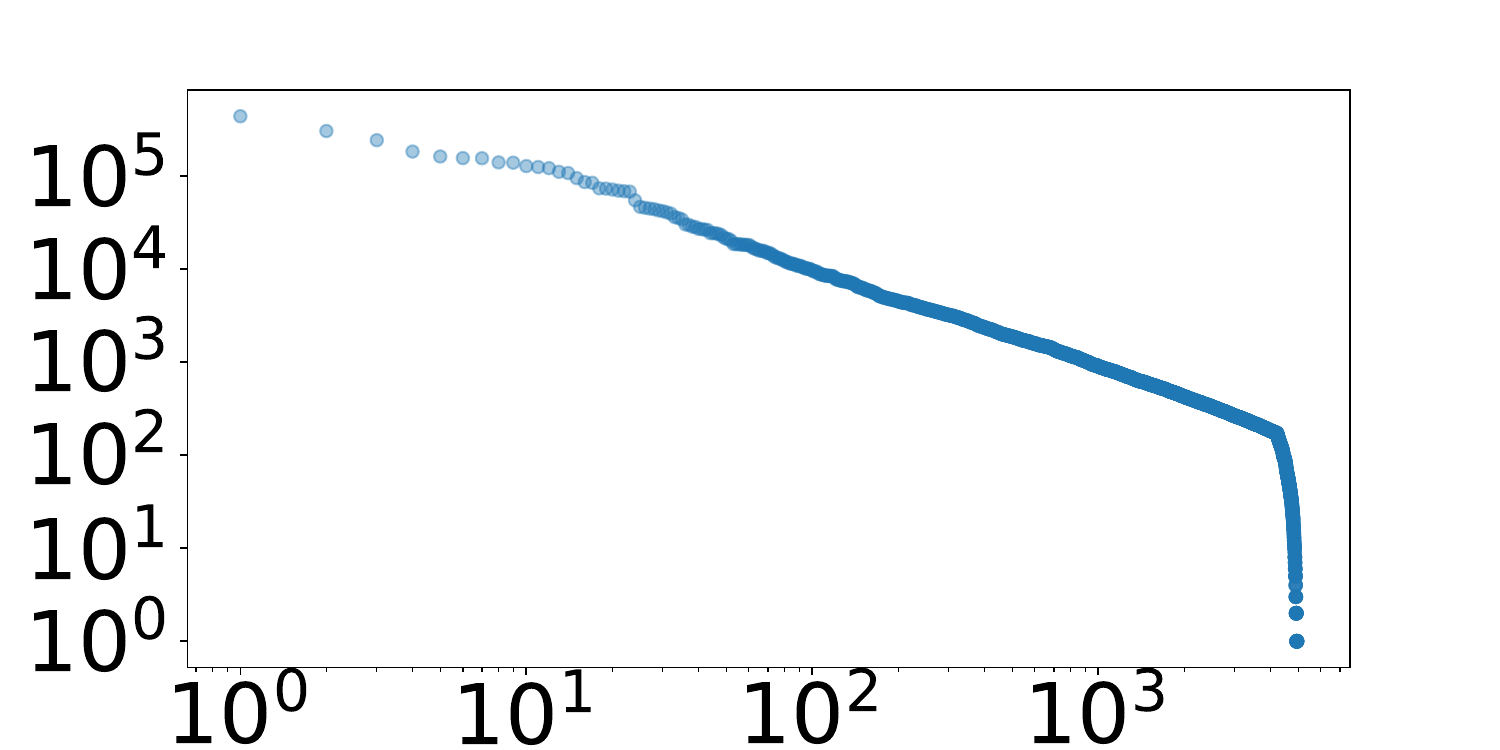}
        \caption{Vocab Size=5000}
        \label{fig:zipf3}
    \end{subfigure}
    \vfill
    \begin{subfigure}[b]{0.32\textwidth}
        \includegraphics[width=\textwidth]{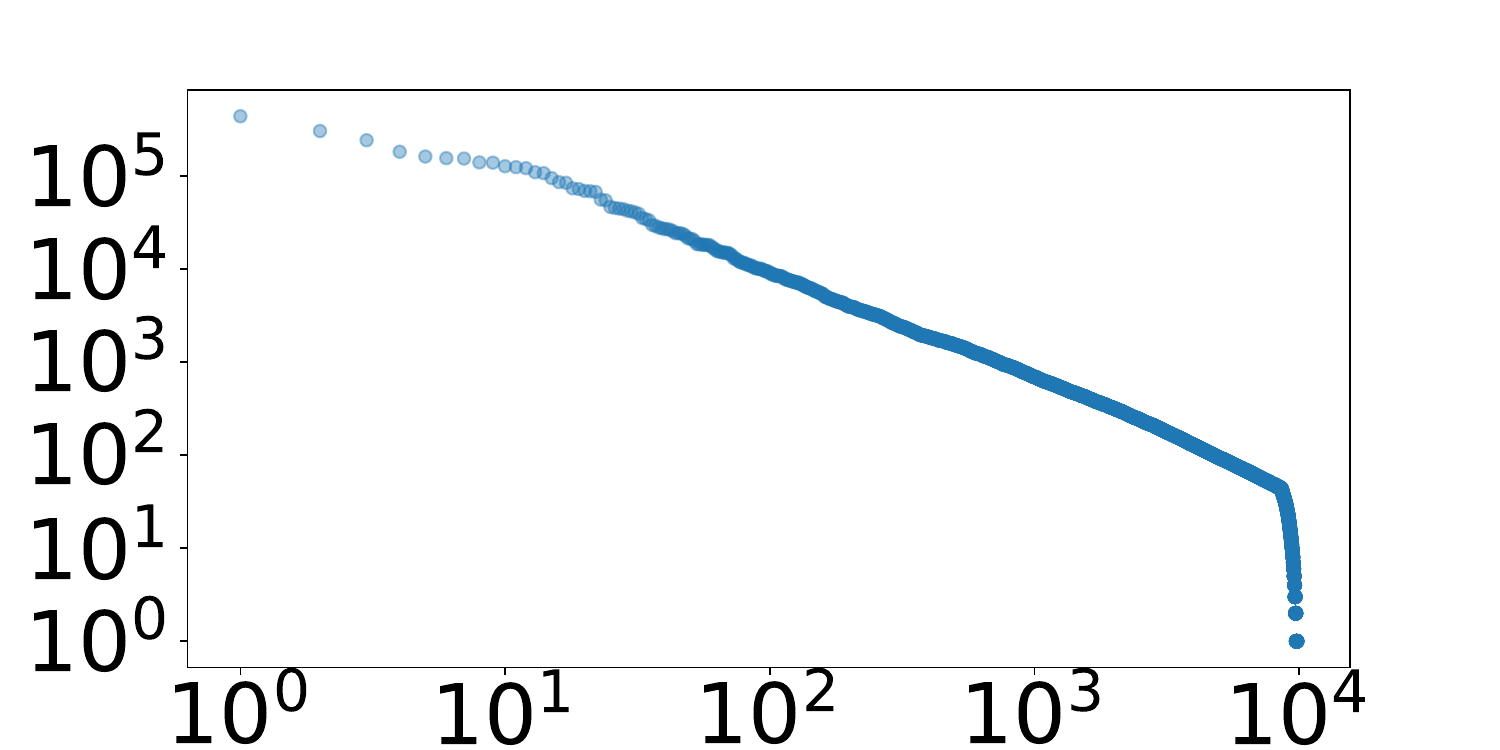}
        \caption{Vocab Size=10000}
        \label{fig:zipf4}
    \end{subfigure}
    \hfill
    \begin{subfigure}[b]{0.32\textwidth}
        \includegraphics[width=\textwidth]{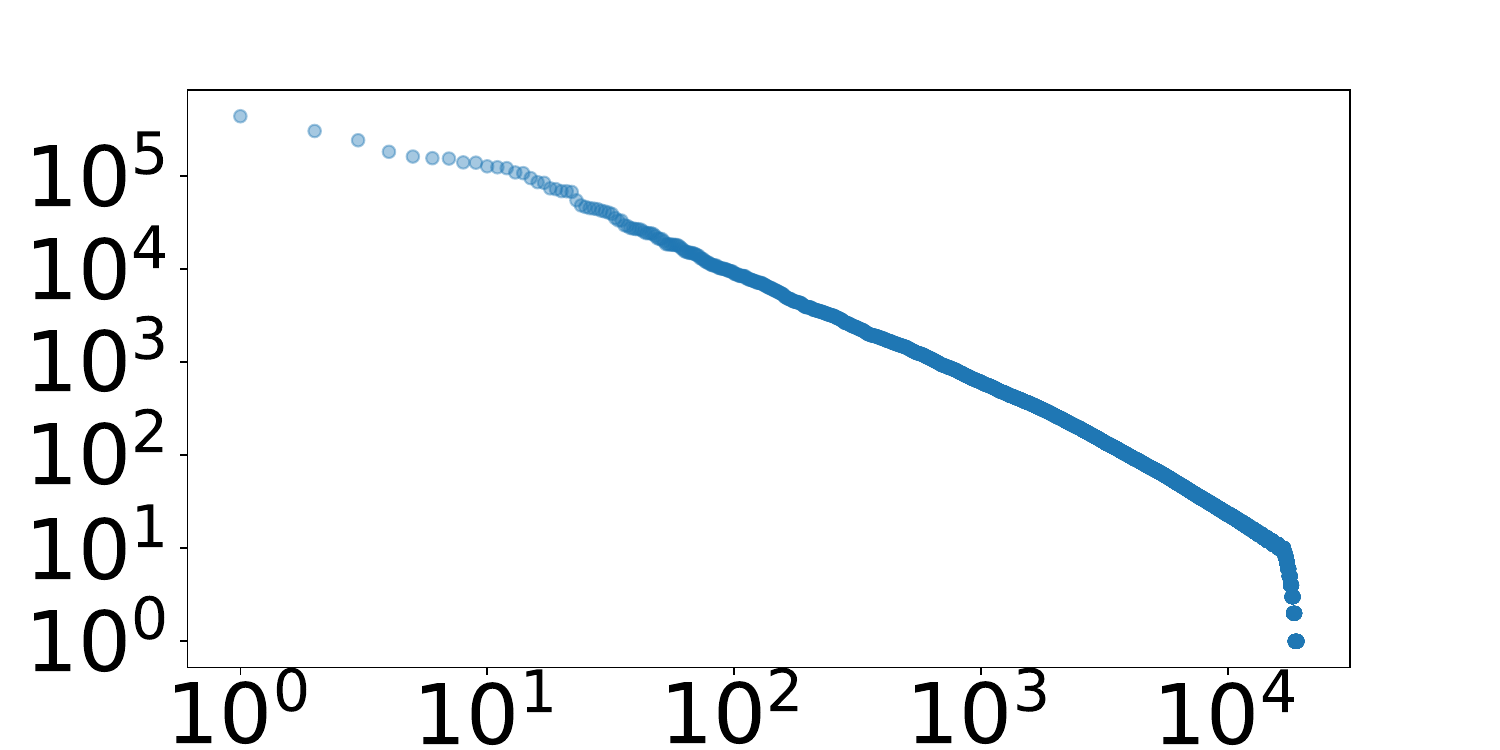}
        \caption{Vocab Size=20000}
        \label{fig:zipf5}
    \end{subfigure}
    \hfill
    \begin{subfigure}[b]{0.32\textwidth}
        \includegraphics[width=\textwidth]{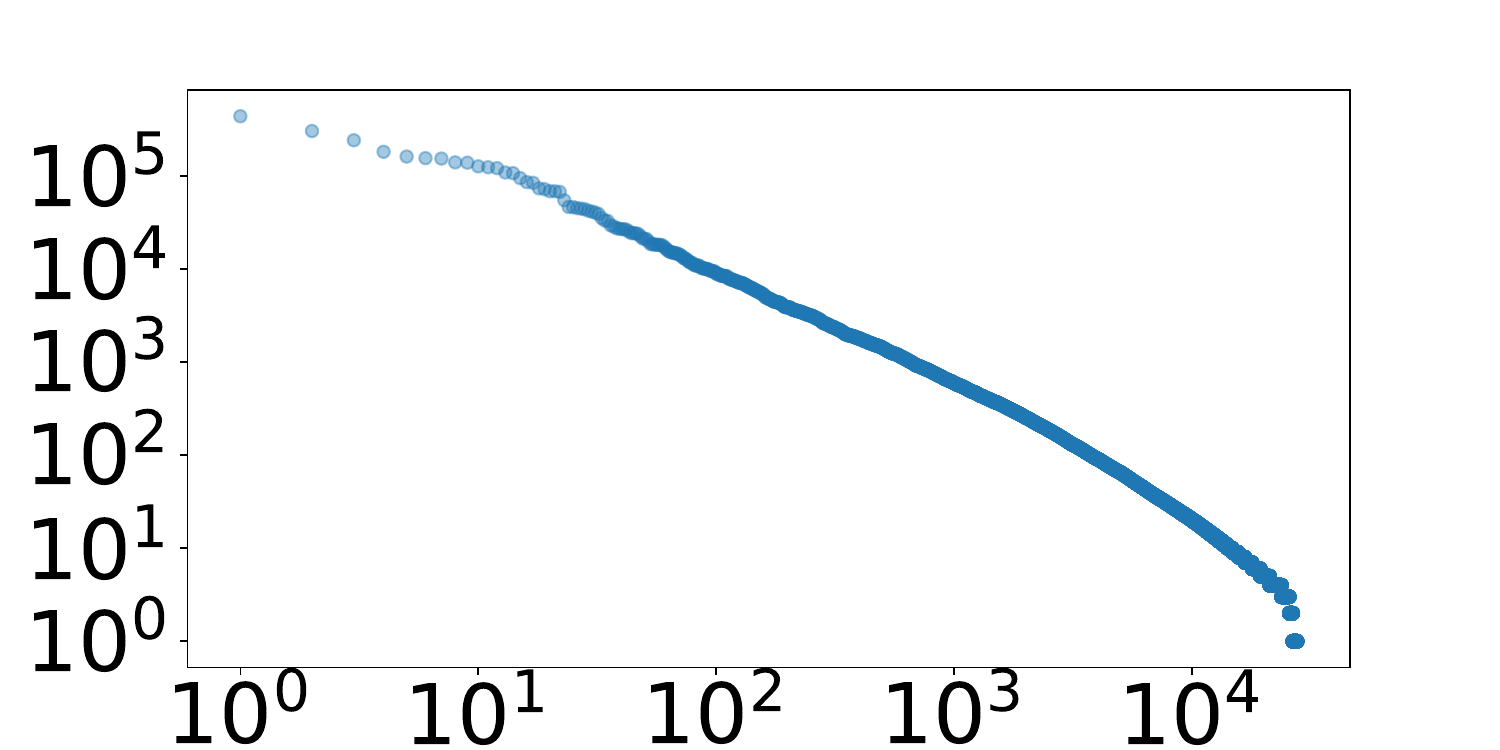}
        \caption{Vocab Size=30000}
        \label{fig:zipf6}
    \end{subfigure}
    \caption{Log-log rank-frequency distribution of different vocabulary sizes on BookCorpus. As the size increases, the curves become increasingly linear, indicating closer adherence to Zipf's law.}
    \label{fig:zipf}
    \vspace{-0.2in}
\end{figure*}

From Figure~\ref{fig:zipf}, we observe the following noteworthy phenomenon:

\textbf{Observation 1:} When the vocabulary is small, the log-log rank-frequency distribution exhibits a clear curvature, deviating significantly from the ideal power-law form. As the vocabulary increases, the curve straightens and approximates a linear trend. This indicates that expanding vocabulary promotes statistical self-organization of token usage, making the token distribution conform more closely to Zipf's law.
This observation directly supports \textbf{Hypothesis 1}, showing that Zipfian behavior emerges naturally as the vocabulary grows. Motivated by this, we design a data-driven vocabulary selection strategy that leverages Zipfian alignment as a stopping criterion for vocabulary expansion.



To automatically determine an appropriate vocabulary size, we design an iterative algorithm that gradually grows the vocabulary and monitors how well the resulting token distribution aligns with Zipf's law. The alignment is quantified using a statistical goodness-of-fit score, such as the coefficient of determination ($R^2$), computed between the empirical log-log rank-frequency curve and an ideal Zipfian distribution.

The procedure begins with a small initial vocabulary and expands it step by step using BPE or a similar merge-based algorithm. After the $t$-th update of vocabulary, we re-tokenize the corpus and calculate the new Zipfian fit score, denoted as $\text{Zipf}_t$. We keep track of the best Zipf score $\text{Zipf}_{\text{max}}$.

To determine when the vocabulary has grown sufficiently, we introduce a stagnation counter that monitors whether further merges lead to meaningful improvements in Zipfian alignment. Specifically, if the score $\text{Zipf}_t$ fails to exceed $\text{Zipf}_{\text{max}}$ by more than a small threshold $\epsilon$ after $N$ steps, we consider the Zipfian fit to have stabilized. At this point, the vocabulary is no longer expanded, and the current vocabulary is taken as the optimal set, denoted by $\mathcal{V}_{\text{opt}}$.

This method adapts vocabulary size to the statistical structure of the data and does not rely on arbitrary preset vocabulary sizes. In Section~\ref{sec:experiments}, we evaluate \textbf{Hypothesis 2} and analyze how Zipfian alignment correlates with downstream task performance across different modalities.

\section{Method}
\label{sec:method}
\subsection{Models and Pre-training Methods}

We conduct experiments across multiple domains, including NLP, genomics (Gene), and chemistry (Chem), to evaluate the impact of vocabulary size on model performance. For each domain, we follow a two-stage process: pre-training on domain-specific datasets and fine-tuning on downstream tasks.
In the NLP domain,  both encoder-only model (e.g., BERT~\cite{devlin2019bert}) and encoder-decoder model (e.g., mBART~\cite{liu2020multilingual}) are evaluated. For BERT, we pre-train the model on a combination of OpenWebText\cite{gokaslan2019openwebtext} and BookCorpus\cite{zhu2015aligning} datasets, following the standard Masked Language Modeling (MLM) objective \cite{devlin2019bert}. The pre-trained BERT model is then fine-tuned  on the GLUE benchmark, which includes tasks such as sentiment analysis, textual entailment, and paraphrase detection, and the  model performance is evaluated using the GLUE score \cite{wang2018glue}.

For mBART, we pre-train the model on the WMT dataset using the Multilingual Denoising Pre-training objective, focusing on three language pairs: German-English (De-En), French-English (Fr-En), and Chinese-English (Zh-En) \cite{liu2020multilingual}. The pre-trained mBART model is fine-tuned on downstream translation tasks for the respective language pairs and the performance is evaluated using the BLEU score \cite{papineni2002bleu}.

In the genomics domain, we follow the approach of DNABERT2 \cite{zhou2024dnabert}, using a BERT-based architecture tailored for DNA sequences. We pre-train the model on DNA sequences from the same dataset used in DNABERT2, employing the MLM objective. Fine-tuning is performed on downstream classification tasks such as promoter prediction and splice site detection, with model performance evaluated using accuracy.

In the chemistry domain, we focus on the SMILES representation of molecular structures, using a BERT-based architecture. We pre-train the model on the first 5 million data in ZINC20, a large dataset of SMILES sequences representing chemical compounds \cite{irwin2005zinc}. Fine-tuning is performed on downstream classification tasks such as molecular property prediction, and performance is evaluated using the ROC-AUC score .

\subsection{Insight for Bigger Models}
Due to resource constraints, we are limited to fine-tuning relatively smaller models. However, \citet{ruder2019transfer} argue that fine-tuning a smaller pre-trained model on a smaller dataset can yield competitive results compared to training a large model from scratch, particularly for specific domains or tasks . Based on this perspective, the conclusions drawn from our experiments on smaller models can also be extended to larger models, offering valuable insights for scaling up model architectures.

\subsection{Finetuning Dataset and Evaluation Metrics}

For NLP tasks, we fine-tune BERT on the GLUE benchmark, excluding the WNLI task \cite{wang2018glue}. The selected tasks and their evaluation metrics are as follows: CoLA uses the Matthews correlation coefficient (MCC); MRPC and QQP use the average of accuracy and F1 score; STS-B uses the average of Pearson and Spearman correlation; and the remaining tasks are evaluated using accuracy.

For NLP tasks with the mBART model, the model is first pre-trained on the WMT dataset \cite{bojar2016findings} for each language pair, and then fine-tuned on the IWSLT dataset, specifically: IWSLT14\cite{cettolo2014report} for De-En, IWSLT17\cite{cettolo2017overview} for Fr-En, and IWSLT15\cite{cettolo-etal-2015-iwslt} for Zh-En.

For genomics tasks with BERT, we use the GUE dataset that has 4 tasks: Core Promoter Detection, Transcription Factor Prediction, Promoter Detection, and Epigenetic Marks Prediction.

For the chemistry tasks with BERT, we use the MoleculeNet dataset, specifically the BBBP, Tox21, Sider, ClinTox, HIV, and BACE datasets, and use ROC-AUC as the evaluation metric .

\subsection{Determining Vocabulary Size}
Determining vocabulary size is crucial for downstream tasks, as different domains require varying levels of token granularity. For NLP tasks, experiments are conducted with BERT vocabulary sizes ranging from 2,000 to 50,000. For multilingual translation tasks, vocab sizes between 2,000 and 140,000 are utilized, as both languages share a common tokenizer. In the genomics and chemistry domains, where the character set is limited, vocab sizes between 500 and 8,000 are employed. This experimental setup enables a systematic analysis of the influence of vocabulary size on model effectiveness across these diverse modalities, providing insights into the optimal tokenizer configurations required for different types of data.

\section{Experiment Results}
\label{sec:experiments}
Building on the empirical foundation established in Section~\ref{sec:zipf_hypothesis}, we now turn to validating \textbf{Hypothesis 2}: that model performance improves when the token rank-frequency distribution closely follows Zipf's law. While Section~\ref{sec:zipf_hypothesis} demonstrated the natural emergence of Zipfian behavior with increasing vocabulary size, this section investigates whether such statistical alignment correlates with improvements in downstream task performance.

To this end, we evaluate the impact of vocabulary size across multiple domains---including \textbf{natural language}, \textbf{genomics}, and \textbf{chemistry}---to test whether Zipfian alignment provides a meaningful criterion for optimizing tokenizer vocabulary. We analyze:

\begin{compactitem}
    \item The relationship between Zipfian goodness-of-fit (measured via $R^2$) and model performance;
    \item How the optimal vocabulary size varies across domains;
    \item Whether alignment with Zipf’s law generalizes beyond NLP to other  modalities;
    \item Case studies and ablations to validate the robustness of our observations.
\end{compactitem}

This analysis provides strong empirical support for using Zipfian properties as an automatic, interpretable, and domain-agnostic guide for vocabulary size selection.

\subsection{Impact on NLP task performance}

\begin{table*}[t]
    \centering
    \small
    \resizebox{\textwidth}{!}{
    \begin{tabular}{lcccccccccc}
        \toprule
        \textbf{Vocab} & \textbf{CoLA} & \textbf{SST-2} & \textbf{MRPC} & \textbf{STS-B} & \textbf{QQP} & \textbf{MNLI} & \textbf{QNLI} & \textbf{RTE} & \textbf{Avg} & \textbf{$R^2$} \\
        \textbf{size} & \textbf{Matthews} & \textbf{Acc.} & \textbf{Acc./F1} & \textbf{Cor.} & \textbf{Acc./F1} & \textbf{Acc.} & \textbf{Acc.} & \textbf{Acc.} &  & \\
        \midrule
        2,000  & 24.83 & 84.64 & 77.49 & 78.46 & 84.63 & 69.74 & 77.31 & 63.52 & 70.08 & 0.6939 \\
        5,000  & 28.87 & 86.07 & 78.41 & 79.42 & 85.78 & 72.03 & 80.71 & 64.22 & 71.94 & 0.7735 \\
        10,000 & 36.02 & 88.78 & 82.54 & 83.62 & 87.37 & 79.01 & 86.25 & 64.57 & 76.02 & 0.8340 \\
        20,000 & 44.22 & 91.61 & 84.33 & 86.79 & 88.92 & 81.42 & 87.74 & 67.32 & 79.04 & 0.8911 \\
        25,000 & 49.73 & 91.25 & 85.63 & 86.82 & 88.97 & 82.03 & 87.91 & 67.54 & 79.99 & 0.9119 \\
        27,500 & 51.79 & 91.84 & 86.02 & 87.14 & 89.25 & 82.21 & 88.34 & 67.89 & 80.56 & 0.9198 \\
        30,000 & \textbf{54.92} & 92.36 & 86.37 & \textbf{87.45} & 89.52 & \textbf{82.52} & \textbf{88.96} & \textbf{68.94} & \textbf{81.38} & 0.9372 \\
        32,500 & 52.37 & \textbf{92.42} & 86.30 & 87.32 & \textbf{89.78} & 82.31 & 88.63 & 68.53 & 80.96 & 0.9344 \\
        35,000 & 53.64 & 92.39 & \textbf{86.42} & 87.42 & 89.63 & 82.51 & 88.72 & 68.76 & 81.19 & 0.9397 \\
        37,500 & 52.97 & 92.47 & 86.29 & 87.21 & 89.54 & 82.34 & 88.52 & 68.69 & 81.00 & 0.9408 \\
        40,000 & 53.27 & 92.21 & 86.24 & 87.37 & 89.31 & 82.29 & 88.65 & 68.42 & 80.97 & 0.9425 \\
        50,000 & 50.23 & 91.83 & 85.95 & 86.47 & 88.88 & 81.94 & 88.26 & 67.94 & 80.19 & 0.9414 \\
        \bottomrule
    \end{tabular}}
    \caption{Performance comparison across various classification tasks. Metrics are accuracy for SST-2, MNLI, QNLI and RTE; Matthews correlation for CoLA; the average of accuracy and F1 scores for MRPC and QQP; and the average of Pearson and Spearman correlations for STS-B. Each configuration is run three times with different random seeds, and the averaged results are taken as the final performance.}
    \label{tab:nlp_results}
    
\end{table*}

To quantify the impact of vocabulary size, we evaluate BERT-Medium models trained with different vocabulary sizes on the GLUE benchmark, covering eight NLP tasks. The results in Table~\ref{tab:nlp_results} indicate that models trained with 30,000 vocabulary size consistently achieve the highest performance. Notably, performance at 30,000 is significantly higher than at smaller vocabulary sizes, while further increasing vocabulary size to 35,000 or 50,000 yields marginal or even slightly worse results.

\begin{figure*}[t]
    \centering
    \begin{subfigure}[b]{0.48\textwidth}
        \includegraphics[width=\textwidth]{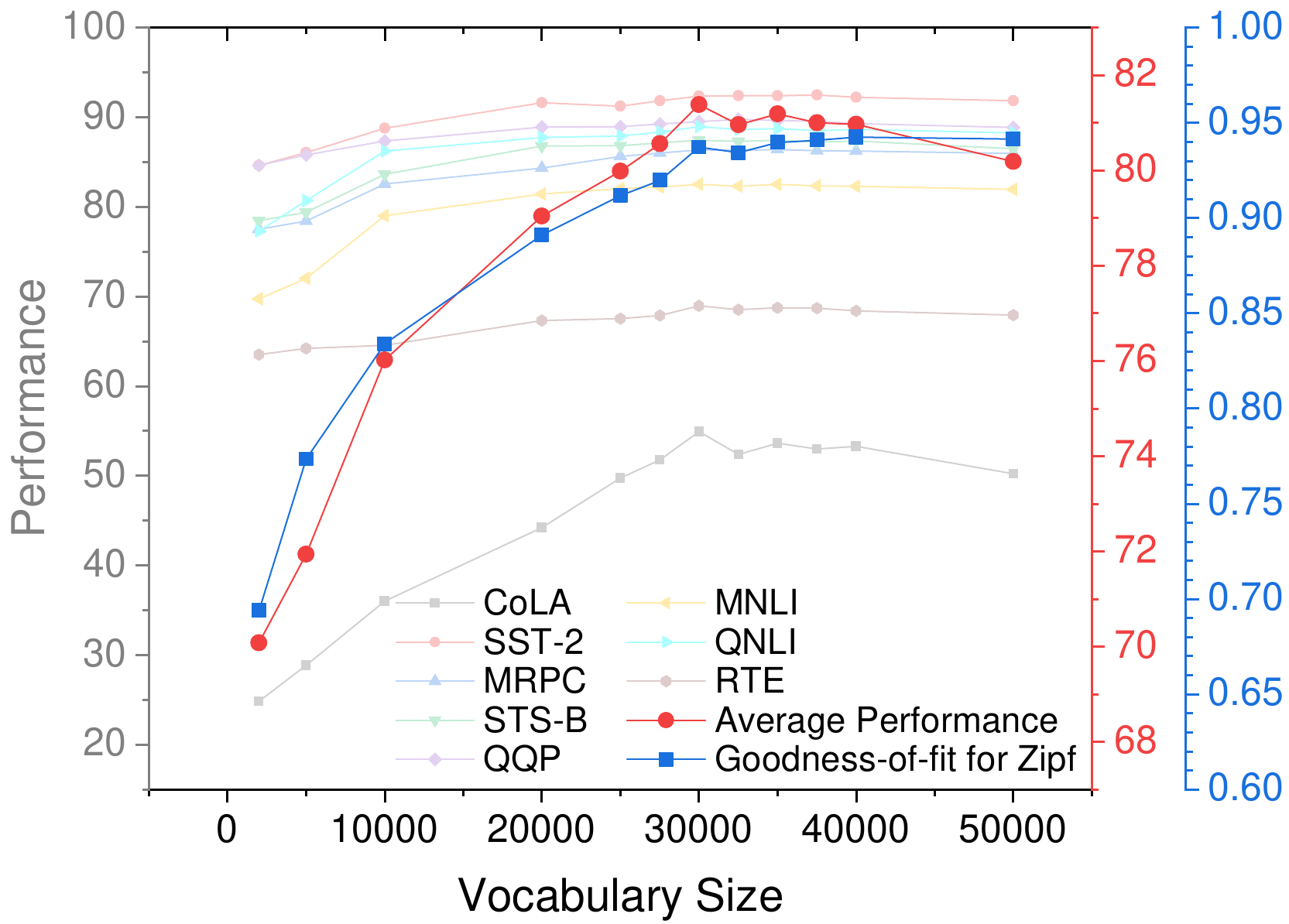}
        \caption{NLP BERT}
        \label{fig:performanceNLP}
    \end{subfigure}
    \hfill 
    \begin{subfigure}[b]{0.48\textwidth}
        \includegraphics[width=\textwidth]{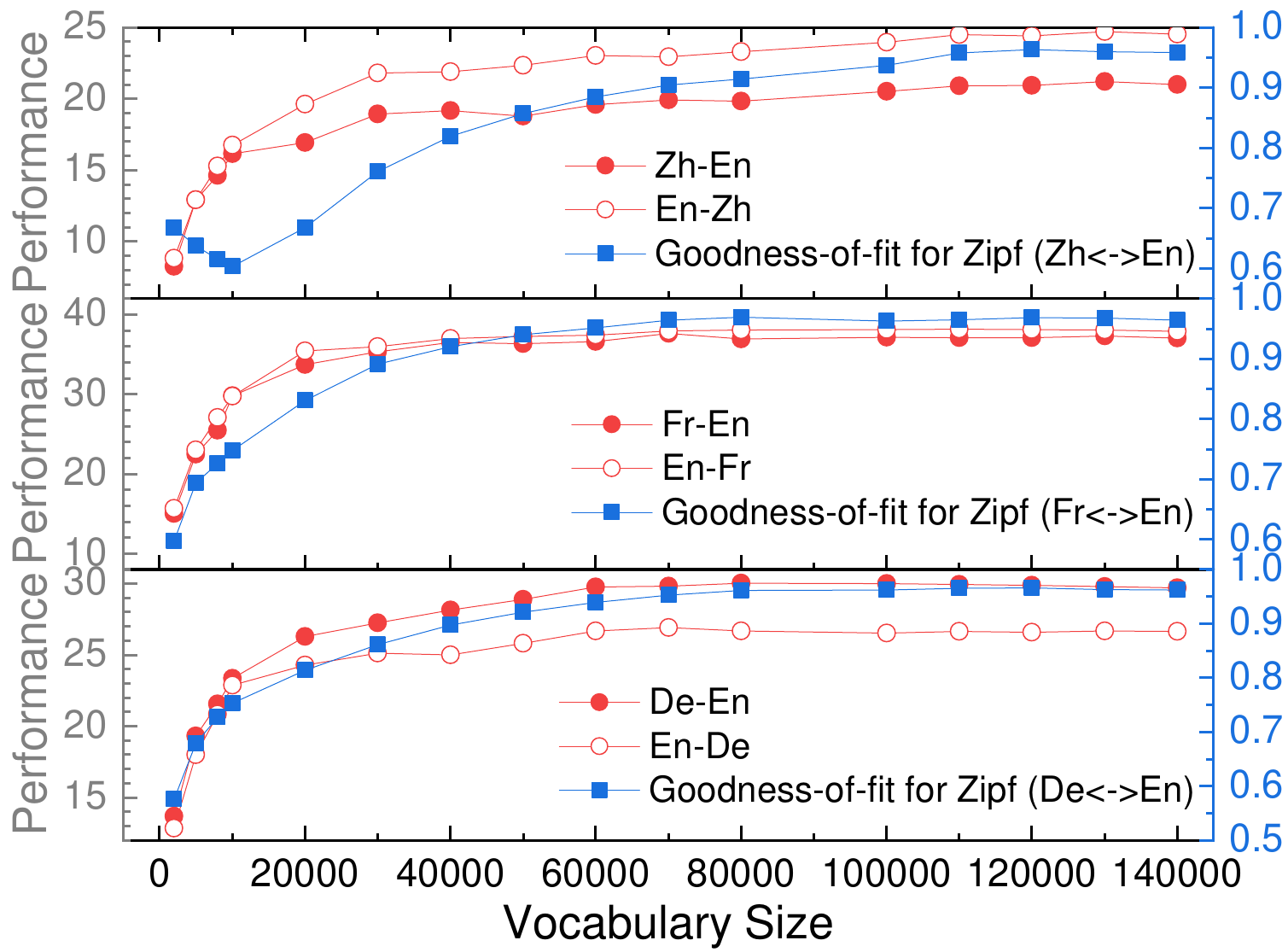}
        \caption{NLP translation}
        \label{fig:performanceTrans}
    \end{subfigure}
    \vfill
    \begin{subfigure}[b]{0.48\textwidth}
        \includegraphics[width=\textwidth]{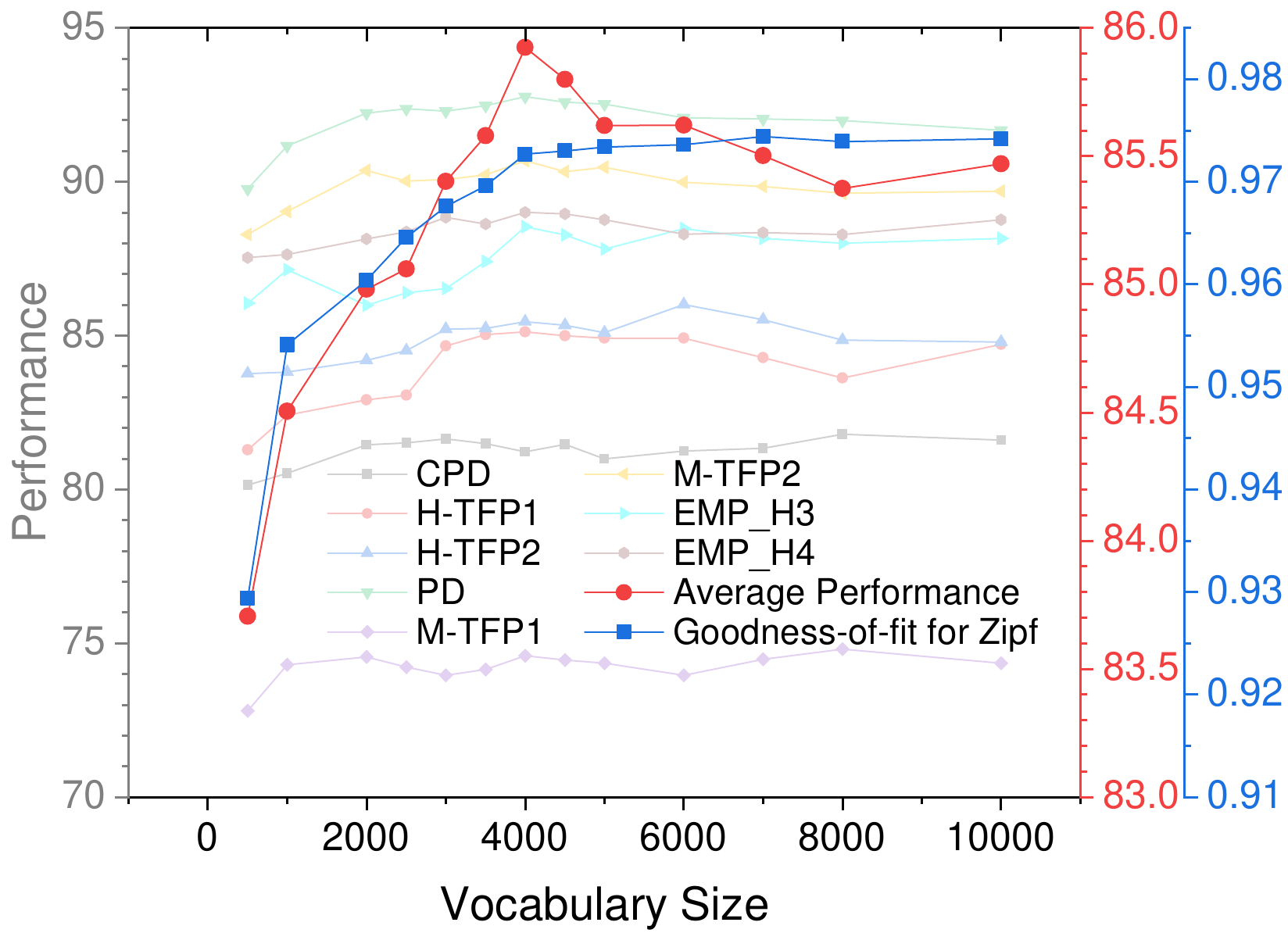}
        \caption{Gene BERT}
        \label{fig:performanceGene}
    \end{subfigure}
    \hfill
    \begin{subfigure}[b]{0.48\textwidth}
        \includegraphics[width=\textwidth]{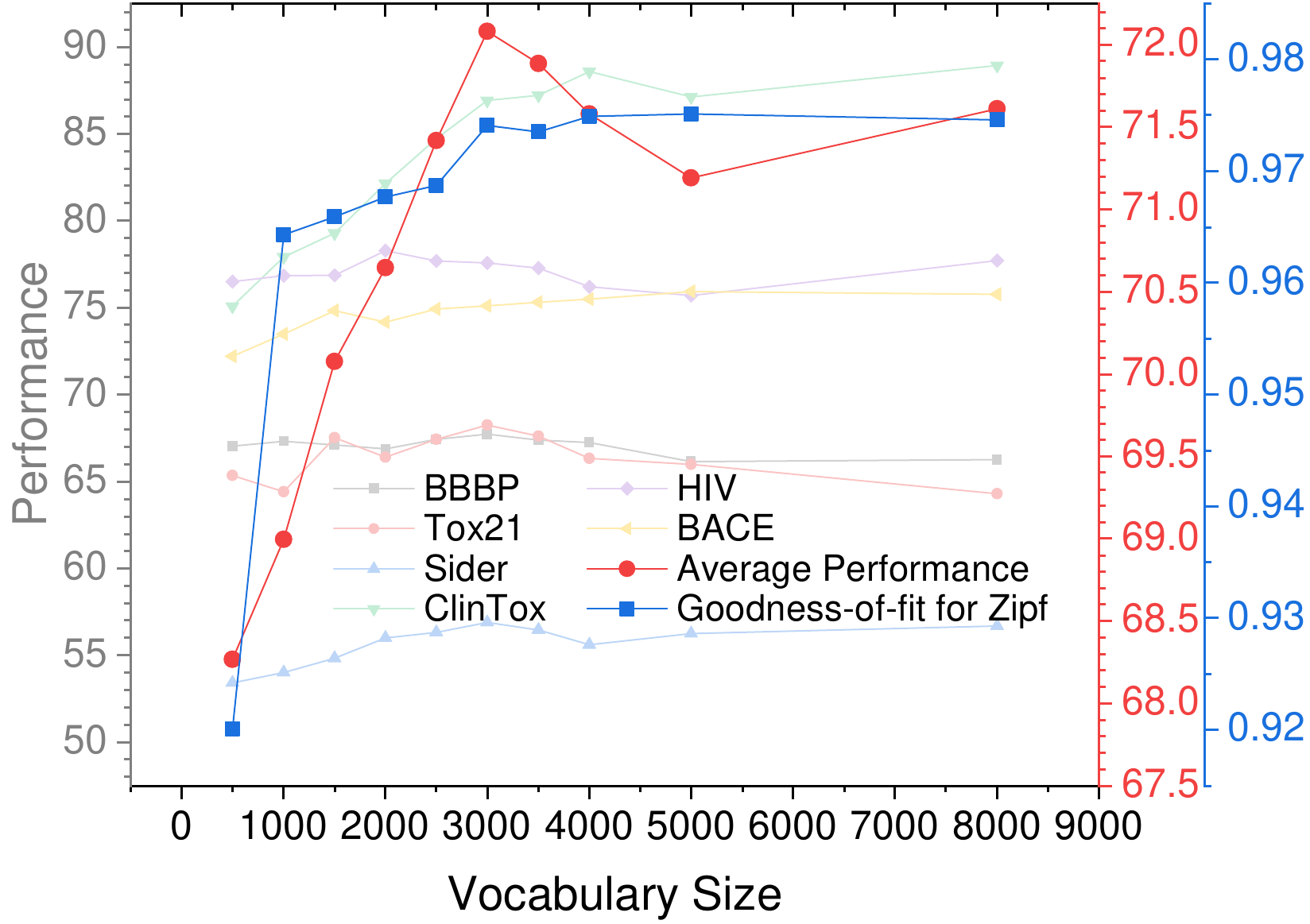}
        \caption{Chem BERT}
        \label{fig:performanceChem}
    \end{subfigure}
    \caption{Model performance with different vocabulary sizes across four distinct domains. Model performance exhibits a consistent trend with the Zipfian goodness-of-fit}
    \label{fig:performance}
    
\end{figure*}
To better illustrate this trend, Figure~\ref{fig:performanceNLP} presents the task performance as a function of vocabulary size. The curve exhibits a clear upward trajectory, peaking at 30,000 before plateauing. \citet{ferrer2001two} empirically demonstrated that word-frequency distributions in large corpora exhibit two distinct power-law regimes, with clear inflection points in the exponent values. This observation motivates the application of segmented fitting and enables a quantitative evaluation of linearity on log-log rank-frequency plots. Alternative validation methods for power-law behavior include maximum-likelihood estimation combined with goodness-of-fit tests based on the Kolmogorov-Smirnov statistic, which measures the greatest vertical deviation between empirical and theoretical cumulative distributions \cite{clauset2009powerlaw}. The Kolmogorov-Smirnov statistic, however, is notably insensitive to variations in the distribution tails, where the most significant power law behavior arises. While providing a more comprehensive assessment by assigning additional weight to tail differences, metrics such as the Kuiper or Anderson-Darling statistics introduce added complexity to the analysis\cite{clauset2009powerlaw}. Given the dual-regime structure observed in Figure~\ref{fig:zipf} and the importance of accurately capturing both the head and the tail of Zipfian distributions, we approximate each rank-frequency distribution with a least squares linear fit and adopt the coefficient of determination \( R^2 \) as goodness-of-fit measure because it offers an intuitive and interpretable quantification of linearity across the entire rank-frequency spectrum.

The results show that as vocabulary size increases, the \( R^2 \) value steadily improves. Specifically, before reaching a vocabulary size of 30,000, the \( R^2 \) value increases rapidly, while after reaching 30,000, the \( R^2 \) value stabilizes at a high value. From Figure~\ref{fig:performanceNLP}, we observe that  \( R^2 \) closely follows the trend of the average performance. This further demonstrates that the closeness to Zipf's law at different vocabulary sizes reflects the performance of downstream tasks.

Similar conclusions can be drawn from the results of the translation tasks (Table~\ref{tab:trans_results} in Appendix ~\ref{sec:appendix2}). When the \( R^2 \) metric reaches its optimal value, the BLEU score is also relatively high. Figure~\ref{fig:performanceChem} illustrates the relationship between the translation task performance and vocabulary size for three language pairs. Obviously, the trend of \(R^2\) is consistent with the task performance.

\textbf{Observation 2:} The token rank-frequency distribution can serve as a prior indicator of a pre-trained model's performance on downstream tasks. When the token distribution approaches a power law, it suggests that the tokenizer is well-suited for the task, leading to better performance on downstream tasks.  This suggests that closeness to Zipf's law can be a useful metric for choosing the best tokenizer and vocabulary.

\subsection{Generalization to Genomics and Chemistry}

To assess its generalizability, the proposed approach is extended to genomics and chemistry, where determining the vocabulary remains an open challenge.

\begin{table*}[t]
    \centering
    \small
    \resizebox{\textwidth}{!}{
    \begin{tabular}{lcccccccccc}
        \toprule
        \textbf{Vocab} & \textbf{CPD} & \textbf{H-TFP1} & \textbf{H-TFP2} & \textbf{PD} & \textbf{M-TFP1} & \textbf{M-TFP2} & \textbf{EMP\_H3} & \textbf{EMP\_H4} & \textbf{Avg} & \textbf{$R^2$} \\
        \textbf{size} & \textbf{Acc.} & \textbf{Acc.} & \textbf{Acc.} & \textbf{Acc.} & \textbf{Acc.} & \textbf{Acc.} & \textbf{Acc.} & \textbf{Acc.} & & \\
        \midrule
        500   & 80.15 & 81.29 & 83.76 & 89.77 & 72.81 & 88.28 & 86.05 & 87.53 & 83.71 & 0.9294 \\
        1,000  & 80.53 & 82.42 & 83.81 & 91.16 & 74.31 & 89.03 & 87.14 & 87.64 & 84.51 & 0.9541 \\
        2,000  & 81.45 & 82.91 & 84.20 & 92.23 & 74.56 & 90.36 & 85.99 & 88.15 & 84.98 & 0.9604 \\
        2,500  & 81.52 & 83.07 & 84.51 & 92.37 & 74.23 & 90.02 & 86.39 & 88.37 & 85.06 & 0.9646 \\
        3,000  & 81.64 & 84.67 & 85.21 & 92.29 & 73.96 & 90.07 & 86.53 & 88.84 & 85.40 & 0.9676 \\
        3,500  & 81.49 & 85.03 & 85.24 & 92.47 & 74.15 & 90.22 & 87.41 & 88.62 & 85.58 & 0.9696 \\
        4,000  & 81.23 & \textbf{85.12} & 85.45 & \textbf{92.76} & 74.60 & \textbf{90.68} & \textbf{88.54} & \textbf{89.01} & \textbf{85.92} & 0.9727 \\
        4,500  & 81.46 & 84.99 & 85.34 & 92.59 & 74.46 & 90.33 & 88.27 & 88.95 & 85.80 & 0.9730 \\
        5,000  & 81.00 & 84.92 & 85.10 & 92.53 & 74.35 & 90.47 & 87.81 & 88.77 & 85.62 & 0.9734 \\
        6,000  & 81.25 & 84.92 & \textbf{86.01} & 92.08 & 73.96 & 89.99 & 88.47 & 88.29 & 85.62 & 0.9736 \\
        7,000  & 81.34 & 84.28 & 85.52 & 92.04 & 74.48 & 89.85 & 88.16 & 88.35 & 85.50 & 0.9744 \\
        8,000  & \textbf{81.80} & 83.62 & 84.85 & 91.99 & \textbf{74.81} & 89.63 & 88.01 & 88.28 & 85.37 & 0.9739 \\
        10,000 & 81.61 & 84.72 & 84.79 & 91.67 & 74.35 & 89.69 & 88.16 & 88.77 & 85.47 & 0.9742 \\
        \bottomrule
    \end{tabular}}
    \caption{Performance comparison of different vocabulary sizes in gene-related classification tasks. Accuracy is reported for all tasks, measuring the performance of BERT-based models on DNA sequence classification. Each configuration is run three times with different random seeds, and the averaged results are reported.}
    \label{tab:gene_results}
    
\end{table*}

In genomics, we pre-train BERT-based models on DNA sequences, following the setup of DNABERT2, and evaluate performance on various GUE classification tasks. The results presented in Table~\ref{tab:gene_results} indicate that optimal performance is achieved with moderate vocabulary sizes, specifically around 4000. Notably, for 5 out of the 8 tasks, the BERT model trained with a 4000-vocabulary-size tokenizer demonstrates superior accuracy scores. As shown in Figure~\ref{fig:performanceGene}, the \( R^2 \) value continues to rise as the vocabulary size increases up to 4000, after which there is no significant improvement. This aligns with our intuition: smaller vocabularies fail to capture biologically meaningful substructures, while excessively large vocabularies lead to redundant segmentations.

\begin{table}[t]
    \hspace*{-0.4cm}
    \centering
    \small
    \renewcommand{\arraystretch}{1.2}  
    \begin{tabular}{p{0.06\linewidth}p{0.06\linewidth}p{0.06\linewidth}p{0.06\linewidth}p{0.07\linewidth}p{0.06\linewidth}p{0.06\linewidth}p{0.06\linewidth}p{0.07\linewidth}}
        \toprule
        \textbf{Vocab} & \fontsize{8pt}{12pt}\textbf{BBBP} & \textbf{Tox21} & \textbf{Sider} & \fontsize{8pt}{12pt}\selectfont\textbf{ClinTox} & \textbf{HIV} & \fontsize{8pt}{12pt}\textbf{BACE} & \textbf{Avg} & \textbf{$R^2$} \\
        \textbf{size} & \textbf{ROC} & \textbf{ROC} & \textbf{ROC} & \textbf{ROC} & \textbf{ROC} & \textbf{ROC} & & \\
        \midrule
        500  & 67.05 & 65.34 & 53.41 & 75.09 & 76.51 & 72.20 & 68.27 & 0.9201 \\
        1,000 & 67.31 & 64.41 & 54.02 & 77.91 & 76.84 & 73.49 & 69.00 & 0.9643 \\
        1500 & 67.12 & 67.51 & 54.83 & 79.30 & 76.87 & 74.83 & 70.08 & 0.9659 \\
        2,000 & 66.89 & 66.39 & 56.00 & 82.14 & \textbf{78.29} & 74.17 & 70.65 & 0.9677 \\
        2,500 & 67.42 & 67.43 & 56.32 & 84.72 & 77.69 & 74.92 & 71.42 & 0.9687 \\
        3,000 & \textbf{67.73} & \textbf{68.26} & \textbf{56.89} & 86.92 & 77.58 & 75.11 & \textbf{72.08} & 0.9741 \\
        3,500 & 67.39 & 67.62 & 56.47 & 87.23 & 77.29 & 75.32 & 71.89 & 0.9735 \\
        4,000 & 67.24 & 66.34 & 55.61 & \textbf{88.59} & 76.20 & 75.50 & 71.58 & 0.9749 \\
        5,000 & 66.14 & 65.99 & 56.26 & 87.12 & 75.70 & \textbf{75.93} & 71.19 & 0.9751 \\
        8,000 & 66.27 & 64.29 & 56.69 & 88.94 & 77.71 & 75.77 & 71.61 & 0.9746 \\
        \bottomrule
    \end{tabular}
    \caption{Performance comparisons are performed on various classification tasks in the MoleculeNet dataset, using ROC-AUC scores as the evaluation metric. Each configuration is run three times with different random seeds, and the average is used as the final performance metric.}
    \vspace{-0.2in}
    \label{tab:chem_results}
\end{table}

Similarly, in chemistry, we tokenize SMILES molecular representations and pre-train models using the ZINC20 dataset. The results presented in Table~\ref{tab:chem_results} indicate that performance continues to improve as vocabulary size increases from 500 to 3000. However, after reaching a vocabulary size of 3000, performance begins to slightly decline with further increases in vocabulary size. A vocabulary size of 3000 yields the best performance, achieving the highest ROC-AUC score and the highest average score. By examining the \( R^2 \) metric in both Table~\ref{tab:chem_results} and Figure~\ref{fig:performanceChem}, we observe that a vocabulary size of 3000 represents the turning point. This finding further supports our Observation 2 in the chemistry domain and provides valuable insight for utilizing an appropriate tokenizer that can effectively capture functional groups in molecular structures.

\subsection{Case Studies: Tokenization Granularity Across Vocabulary Sizes}
In the case studies section, we provide examples to show that having a vocabulary that is too small or too large is not appropriate. The figure shows examples from both the NLP and chemistry domains to illustrate this conclusion.

In the first example below, we do analysis for \texttt{CCCOc1ccc(cc1)c2cccc3c2nccn3} -- the SMILES representation of the molecule, and compare how different vocabulary sizes affect its tokenization. With a small vocabulary size, the molecule is overly fragmented—for instance, into tokens like \texttt{c1ccc} and \texttt{ccn3(cc1)}—which breaks apart chemically meaningful structures and leads to unstable or uninterpretable fragments. At an appropriate vocabulary size, the tokenizer produces segments such as \texttt{CCCOc1ccc}, \texttt{(cc1)}, and \texttt{c2cccc3c2}, which aligns with functional groups and aromatic or heterocyclic rings, enhancing chemical interpretability. However, when the vocabulary is too large, tokens like \texttt{c2cccc3c2n} emerge, which over-merge frequent but semantically inconsistent character sequences. These tokens span across distinct substructures, disrupting meaningful chemical units and weakening the tokenizer’s ability to preserve domain-relevant structure. This observation reinforces the importance of choosing a vocabulary size that balances token compactness with chemical coherence.

In the second example, we show how the phrase ``invisible footprints'' is tokenized with a vocabulary size of 30,000, correctly splitting it into ``in'' ``visible'' ``foot'' ``prints''. When a smaller vocabulary size is used, the word is split into non-semantic tokens such as ``in'' ``vis''  ``ible'' ``foot'' ``prin'' ``ts'' resulting in a loss of semantic meaning. When the vocabulary size is too large, each word is treated as a single token, introducing semantic redundancy.

\begin{figure}[ht]
    \centering
    \includegraphics[width=0.5\textwidth]{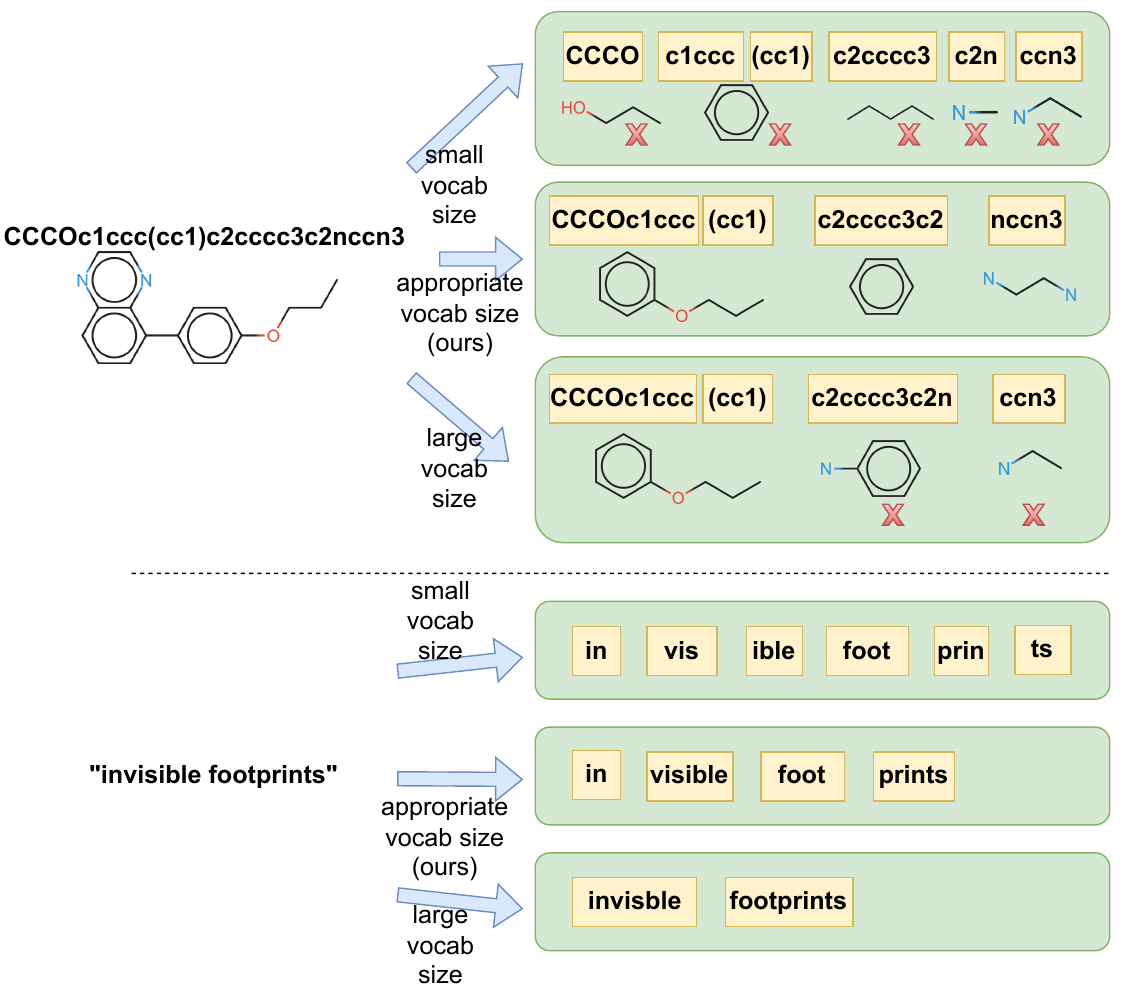}
    \caption{Case study: With an appropriate vocabulary size, the tokenization not only is more effective but also captures essential patterns of sequences.}
    \label{fig:case}
    \vspace{-0.2in}
\end{figure}

These examples further support our approach, providing insights into how vocabulary size influences tokenization quality and, in turn, impacts task performance. They reinforce our method that vocabulary size determining should be Zipfian-guided, ensuring that tokenization reflects intrinsic linguistic and structural patterns.

\section{Conclusion}
\label{sec:conclusion}

This study explored the impact of tokenizer vocabulary size on the performance of pre-trained language models across various domains, including natural language processing, genomics, and chemistry. By analyzing the relationship between token rank-frequency distribution and task performance, we demonstrated that aligning token distributions with power-law scaling laws can serve as a robust criterion for determining optimal vocabulary sizes. Our experiments revealed that models achieve superior performance when the token distribution closely adheres to Zipf's law, indicating that this alignment enhances both efficiency and effectiveness in downstream tasks.

\section*{Acknowledgements}
This work was supported by NSF IIS-2119531,
IIS-2137396, IIS-2142827, IIS-2234058, CCF-1901059, and ONR N00014-22-1-2507.

\section{Limitations}
While our study provides valuable insights into the relationship between tokenizer vocabulary size and model performance, several limitations should be acknowledged.

Due to hardware limitations, we only conduct pre-training experiments on relatively small models. Although the conclusions drawn from these smaller models offer meaningful guidance for larger models, the significant difference in parameter scale means that our findings may not fully generalize to state-of-the-art architectures with billions of parameters. Further experiments on larger models are necessary to solidify our conclusions and validate the scalability of our approach.

Our experiments primarily focused on a subset of modalities (e.g., NLP, genomics, and chemistry) and a limited range of pre-trained model architectures (e.g., BERT and mBART). To further generalize our findings, future work should extend the evaluation to additional modalities (e.g., vision, audio) and diverse model architectures (e.g., Transformer variants, hybrid models).

\bibliography{custom}

\begin{thebibliography}{36}
\providecommand{\natexlab}[1]{#1}

\bibitem[{Achiam et~al.(2023)Achiam, Adler, Agarwal, Ahmad, Akkaya, Aleman, Almeida, Altenschmidt, Altman, Anadkat et~al.}]{achiam2023gpt}
Josh Achiam, Steven Adler, Sandhini Agarwal, Lama Ahmad, Ilge Akkaya, Florencia~Leoni Aleman, Diogo Almeida, Janko Altenschmidt, Sam Altman, Shyamal Anadkat, and 1 others. 2023.
\newblock Gpt-4 technical report.
\newblock \emph{arXiv preprint arXiv:2303.08774}.

\bibitem[{Ali et~al.(2024)Ali, Fromm, Thellmann, Rutmann, L{\"u}bbering, Leveling, Klug, Ebert, Doll, Buschhoff et~al.}]{ali2024tokenizer}
Mehdi Ali, Michael Fromm, Klaudia Thellmann, Richard Rutmann, Max L{\"u}bbering, Johannes Leveling, Katrin Klug, Jan Ebert, Niclas Doll, Jasper Buschhoff, and 1 others. 2024.
\newblock Tokenizer choice for {LLM} training: Negligible or crucial?
\newblock In \emph{Findings of the Association for Computational Linguistics: NAACL 2024}, pages 3907--3924.

\bibitem[{Barab{\'a}si and Albert(1999)}]{barabasi1999emergence}
Albert-L{\'a}szl{\'o} Barab{\'a}si and R{\'e}ka Albert. 1999.
\newblock Emergence of scaling in random networks.
\newblock \emph{science}, 286(5439):509--512.

\bibitem[{Bojar et~al.(2016)Bojar, Chatterjee, Federmann, Graham, Haddow, Huck, Yepes, Koehn, Logacheva, Monz et~al.}]{bojar2016findings}
Ondrej Bojar, Rajen Chatterjee, Christian Federmann, Yvette Graham, Barry Haddow, Matthias Huck, Antonio~Jimeno Yepes, Philipp Koehn, Varvara Logacheva, Christof Monz, and 1 others. 2016.
\newblock Findings of the 2016 conference on machine translation ({WMT16}).
\newblock In \emph{First conference on machine translation}, pages 131--198. Association for Computational Linguistics.

\bibitem[{Brown et~al.(2020)Brown, Mann, Ryder, Subbiah, Kaplan, Dhariwal, Neelakantan, Shyam, Sastry, Askell et~al.}]{brown2020language}
Tom Brown, Benjamin Mann, Nick Ryder, Melanie Subbiah, Jared~D Kaplan, Prafulla Dhariwal, Arvind Neelakantan, Pranav Shyam, Girish Sastry, Amanda Askell, and 1 others. 2020.
\newblock Language models are few-shot learners.
\newblock \emph{Advances in neural information processing systems}, 33:1877--1901.

\bibitem[{Cancho and Solé(2001)}]{ferrer2001two}
Ramon~Ferrer Cancho and Ricard~V. Solé. 2001.
\newblock \href {https://doi.org/10.1076/jqul.8.3.165.4101} {Two regimes in the frequency of words and the origins of complex lexicons: Zipf's law revisited}.
\newblock \emph{Journal of Quantitative Linguistics}, 8(3):165--173.

\bibitem[{Cettolo et~al.(2017)Cettolo, Federico, Bentivogli, Niehues, St{\"u}ker, Sudoh, Yoshino, and Federmann}]{cettolo2017overview}
Mauro Cettolo, Marcello Federico, Luisa Bentivogli, Jan Niehues, Sebastian St{\"u}ker, Katsuitho Sudoh, Koichiro Yoshino, and Christian Federmann. 2017.
\newblock Overview of the iwslt 2017 evaluation campaign.
\newblock In \emph{Proceedings of the 14th International Workshop on Spoken Language Translation}, pages 2--14.

\bibitem[{Cettolo et~al.(2015)Cettolo, Niehues, St{\"u}ker, Bentivogli, Cattoni, and Federico}]{cettolo-etal-2015-iwslt}
Mauro Cettolo, Jan Niehues, Sebastian St{\"u}ker, Luisa Bentivogli, Roldano Cattoni, and Marcello Federico. 2015.
\newblock \href {https://aclanthology.org/2015.iwslt-evaluation.1/} {The {IWSLT} 2015 evaluation campaign}.
\newblock In \emph{Proceedings of the 12th International Workshop on Spoken Language Translation: Evaluation Campaign}, pages 2--14, Da Nang, Vietnam.

\bibitem[{Cettolo et~al.(2014)Cettolo, Niehues, St{\"u}ker, Bentivogli, and Federico}]{cettolo2014report}
Mauro Cettolo, Jan Niehues, Sebastian St{\"u}ker, Luisa Bentivogli, and Marcello Federico. 2014.
\newblock Report on the 11th iwslt evaluation campaign.
\newblock In \emph{Proceedings of the 11th International Workshop on Spoken Language Translation: Evaluation Campaign}, pages 2--17.

\bibitem[{Clauset et~al.(2009)Clauset, Shalizi, and Newman}]{clauset2009powerlaw}
Aaron Clauset, Cosma~Rohilla Shalizi, and M.~E.~J. Newman. 2009.
\newblock \href {https://doi.org/10.1137/070710111} {Power-law distributions in empirical data}.
\newblock \emph{{SIAM} Review}, 51(4):661--703.

\bibitem[{Devlin et~al.(2019)Devlin, Chang, Lee, and Toutanova}]{devlin2019bert}
Jacob Devlin, Ming-Wei Chang, Kenton Lee, and Kristina Toutanova. 2019.
\newblock {BERT}: Pre-training of deep bidirectional transformers for language understanding.
\newblock In \emph{Proceedings of the 2019 conference of the North {American} chapter of the association for computational linguistics: human language technologies, volume 1 (long and short papers)}, pages 4171--4186.

\bibitem[{Dosovitskiy et~al.(2021)Dosovitskiy, Beyer, Kolesnikov, Weissenborn, Zhai, Unterthiner, Dehghani, Minderer, Heigold, Gelly, Uszkoreit, and Houlsby}]{dosovitskiy2021image}
Alexey Dosovitskiy, Lucas Beyer, Alexander Kolesnikov, Dirk Weissenborn, Xiaohua Zhai, Thomas Unterthiner, Mostafa Dehghani, Matthias Minderer, Georg Heigold, Sylvain Gelly, Jakob Uszkoreit, and Neil Houlsby. 2021.
\newblock \href {https://openreview.net/forum?id=YicbFdNTTy} {An image is worth 16x16 words: Transformers for image recognition at scale}.
\newblock In \emph{International Conference on Learning Representations}.

\bibitem[{Gage(1994)}]{gage1994new}
Philip Gage. 1994.
\newblock A new algorithm for data compression.
\newblock \emph{The C Users Journal}, 12(2):23--38.

\bibitem[{Gokaslan and Cohen(2019)}]{gokaslan2019openwebtext}
Aaron Gokaslan and Vanya Cohen. 2019.
\newblock Openwebtext corpus.
\newblock \url{https://skylion007.github.io/OpenWebTextCorpus/}.
\newblock Accessed: 2024-05-20.

\bibitem[{Goldman et~al.(2024)Goldman, Caciularu, Eyal, Cao, Szpektor, and Tsarfaty}]{goldman2024unpacking}
Omer Goldman, Avi Caciularu, Matan Eyal, Kris Cao, Idan Szpektor, and Reut Tsarfaty. 2024.
\newblock \href {https://doi.org/10.18653/v1/2024.findings-acl.134} {Unpacking tokenization: Evaluating text compression and its correlation with model performance}.
\newblock In \emph{Findings of the Association for Computational Linguistics: ACL 2024}, pages 2274--2286, Bangkok, Thailand. Association for Computational Linguistics.

\bibitem[{Irwin and Shoichet(2005)}]{irwin2005zinc}
John~J Irwin and Brian~K Shoichet. 2005.
\newblock {ZINC}--a free database of commercially available compounds for virtual screening.
\newblock \emph{Journal of chemical information and modeling}, 45(1):177--182.

\bibitem[{Jeong et~al.(2001)Jeong, Mason, Barab{\'a}si, and Oltvai}]{jeong2001lethality}
Hawoong Jeong, Sean~P Mason, A-L Barab{\'a}si, and Zoltan~N Oltvai. 2001.
\newblock Lethality and centrality in protein networks.
\newblock \emph{Nature}, 411(6833):41--42.

\bibitem[{Ji et~al.(2021)Ji, Zhou, Liu, and Davuluri}]{ji2021dnabert}
Yanrong Ji, Zhihan Zhou, Han Liu, and Ramana~V Davuluri. 2021.
\newblock {DNABERT}: pre-trained bidirectional encoder representations from transformers model for {DNA}-language in genome.
\newblock \emph{Bioinformatics}, 37(15):2112--2120.

\bibitem[{Kudo and Richardson(2018)}]{kudo2018sentencepiece}
Taku Kudo and John Richardson. 2018.
\newblock \href {https://doi.org/10.18653/v1/D18-2012} {{S}entence{P}iece: A simple and language independent subword tokenizer and detokenizer for neural text processing}.
\newblock In \emph{Proceedings of the 2018 Conference on Empirical Methods in Natural Language Processing: System Demonstrations}, pages 66--71, Brussels, Belgium. Association for Computational Linguistics.

\bibitem[{Liu et~al.(2020)Liu, Gu, Goyal, Li, Edunov, Ghazvininejad, Lewis, and Zettlemoyer}]{liu2020multilingual}
Yinhan Liu, Jiatao Gu, Naman Goyal, Xian Li, Sergey Edunov, Marjan Ghazvininejad, Mike Lewis, and Luke Zettlemoyer. 2020.
\newblock \href {https://doi.org/10.1162/tacl_a_00343} {Multilingual denoising pre-training for neural machine translation}.
\newblock \emph{Transactions of the Association for Computational Linguistics}, 8:726--742.

\bibitem[{Montemurro(2001)}]{montemurro2001beyond}
Marcelo~A Montemurro. 2001.
\newblock Beyond the {Zipf}--{Mandelbrot} law in quantitative linguistics.
\newblock \emph{Physica A: Statistical Mechanics and its Applications}, 300(3-4):567--578.

\bibitem[{Papineni et~al.(2002)Papineni, Roukos, Ward, and Zhu}]{papineni2002bleu}
Kishore Papineni, Salim Roukos, Todd Ward, and Wei-Jing Zhu. 2002.
\newblock \href {https://doi.org/10.3115/1073083.1073135} {{BLEU}: a method for automatic evaluation of machine translation}.
\newblock In \emph{Proceedings of the 40th Annual Meeting of the Association for Computational Linguistics}, pages 311--318, Philadelphia, Pennsylvania, USA. Association for Computational Linguistics.

\bibitem[{Pareto(1964)}]{pareto1964cours}
Vilfredo Pareto. 1964.
\newblock \emph{Cours d'{\'e}conomie politique}, volume~1.
\newblock Librairie Droz.

\bibitem[{Powers(1998)}]{powers1998applications}
David M.~W. Powers. 1998.
\newblock Applications and explanations of zipf's law.
\newblock In \emph{Proceedings of the Joint Conferences on New Methods in Language Processing and Computational Natural Language Learning}, NeMLaP3/CoNLL '98, page 151–160, USA. Association for Computational Linguistics.

\bibitem[{Provilkov et~al.(2020)Provilkov, Emelianenko, and Voita}]{provilkov2020bpe}
Ivan Provilkov, Dmitrii Emelianenko, and Elena Voita. 2020.
\newblock \href {https://doi.org/10.18653/v1/2020.acl-main.170} {{BPE}-dropout: Simple and effective subword regularization}.
\newblock In \emph{Proceedings of the 58th Annual Meeting of the Association for Computational Linguistics}, pages 1882--1892, Online. Association for Computational Linguistics.

\bibitem[{Radford et~al.(2023)Radford, Kim, Xu, Brockman, McLeavey, and Sutskever}]{radford2023robust}
Alec Radford, Jong~Wook Kim, Tao Xu, Greg Brockman, Christine McLeavey, and Ilya Sutskever. 2023.
\newblock Robust speech recognition via large-scale weak supervision.
\newblock In \emph{Proceedings of the 40th International Conference on Machine Learning}, ICML'23. PMLR.

\bibitem[{Radford et~al.(2019)Radford, Wu, Child, Luan, Amodei, Sutskever et~al.}]{radford2019language}
Alec Radford, Jeffrey Wu, Rewon Child, David Luan, Dario Amodei, Ilya Sutskever, and 1 others. 2019.
\newblock Language models are unsupervised multitask learners.
\newblock \emph{OpenAI blog}.

\bibitem[{Ruder et~al.(2019)Ruder, Peters, Swayamdipta, and Wolf}]{ruder2019transfer}
Sebastian Ruder, Matthew~E. Peters, Swabha Swayamdipta, and Thomas Wolf. 2019.
\newblock \href {https://doi.org/10.18653/v1/N19-5004} {Transfer learning in natural language processing}.
\newblock In \emph{Proceedings of the 2019 Conference of the North {A}merican Chapter of the Association for Computational Linguistics: Tutorials}, pages 15--18, Minneapolis, Minnesota. Association for Computational Linguistics.

\bibitem[{Schuster and Nakajima(2012)}]{schuster2012japanese}
Mike Schuster and Kaisuke Nakajima. 2012.
\newblock \href {https://doi.org/10.1109/ICASSP.2012.6289079} {Japanese and korean voice search}.
\newblock In \emph{2012 IEEE International Conference on Acoustics, Speech and Signal Processing (ICASSP)}, pages 5149--5152.

\bibitem[{Schwaller et~al.(2019)Schwaller, Laino, Gaudin, Bolgar, Hunter, Bekas, and Lee}]{schwaller2019molecular}
Philippe Schwaller, Teodoro Laino, Th{\'e}ophile Gaudin, Peter Bolgar, Christopher~A. Hunter, Costas Bekas, and Alpha~A. Lee. 2019.
\newblock \href {https://doi.org/10.1021/acscentsci.9b00576} {Molecular transformer: A model for uncertainty-calibrated chemical reaction prediction}.
\newblock \emph{ACS Central Science}, 5(9):1572--1583.

\bibitem[{Sennrich et~al.(2016)Sennrich, Haddow, and Birch}]{sennrich2016neural}
Rico Sennrich, Barry Haddow, and Alexandra Birch. 2016.
\newblock \href {https://doi.org/10.18653/v1/P16-1162} {Neural machine translation of rare words with subword units}.
\newblock In \emph{Proceedings of the 54th Annual Meeting of the Association for Computational Linguistics (Volume 1: Long Papers)}, pages 1715--1725, Berlin, Germany. Association for Computational Linguistics.

\bibitem[{Wang et~al.(2018)Wang, Singh, Michael, Hill, Levy, and Bowman}]{wang2018glue}
Alex Wang, Amanpreet Singh, Julian Michael, Felix Hill, Omer Levy, and Samuel Bowman. 2018.
\newblock \href {https://doi.org/10.18653/v1/W18-5446} {{GLUE}: A multi-task benchmark and analysis platform for natural language understanding}.
\newblock In \emph{Proceedings of the 2018 {EMNLP} Workshop {B}lackbox{NLP}: Analyzing and Interpreting Neural Networks for {NLP}}, pages 353--355, Brussels, Belgium. Association for Computational Linguistics.

\bibitem[{Wu et~al.(2016)Wu, Schuster, Chen, Le, Norouzi, Macherey, Krikun, Cao, Gao, Macherey et~al.}]{wu2016google}
Yonghui Wu, Mike Schuster, Zhifeng Chen, Quoc~V Le, Mohammad Norouzi, Wolfgang Macherey, Maxim Krikun, Yuan Cao, Qin Gao, Klaus Macherey, and 1 others. 2016.
\newblock Google's neural machine translation system: Bridging the gap between human and machine translation.
\newblock \emph{arXiv preprint arXiv:1609.08144}.

\bibitem[{Zhou et~al.(2024)Zhou, Ji, Li, Dutta, Davuluri, and Liu}]{zhou2024dnabert}
Zhihan Zhou, Yanrong Ji, Weijian Li, Pratik Dutta, Ramana~V Davuluri, and Han Liu. 2024.
\newblock \href {https://openreview.net/forum?id=oMLQB4EZE1} {{DNABERT}-2: Efficient foundation model and benchmark for multi-species genomes}.
\newblock In \emph{The Twelfth International Conference on Learning Representations}.

\bibitem[{Zhu et~al.(2015)Zhu, Kiros, Zemel, Salakhutdinov, Urtasun, Torralba, and Fidler}]{zhu2015aligning}
Yukun Zhu, Ryan Kiros, Richard~S Zemel, Ruslan Salakhutdinov, Raquel Urtasun, Antonio Torralba, and Sanja Fidler. 2015.
\newblock Aligning books and movies: Towards story-like visual explanations by watching movies and reading books.
\newblock In \emph{Proceedings of the IEEE International Conference on Computer Vision (ICCV)}, pages 19--27.

\bibitem[{Zipf(2013)}]{zipf2013psycho}
George~Kingsley Zipf. 2013.
\newblock \emph{The psycho-biology of language: An introduction to dynamic philology}.
\newblock Routledge.

\end{thebibliography}

\appendix
\section{BPE algorithm}
\label{sec:appendix1}
This shows a detailed description of BPE algorithm.

\begin{algorithm}[H]
    \caption{Byte Pair Encoding (BPE)}
    \label{alg:bpe}
    \begin{algorithmic}[1]
        \Require Corpus $D$, target vocabulary size $V$
        \Ensure Vocabulary set $\mathcal{V}$
        \State Initialize $\mathcal{V}$ with all unique characters in $D$ 
        \State Compute frequency of all adjacent symbol pairs in $D$
        \While{$|\mathcal{V}| < V$} \Comment{Continue until target vocabulary size is reached}
            \State Identify the most frequent pair $(s_i, s_j)$ in $D$
            \State Merge $(s_i, s_j)$ into a new symbol $s_k$ 
            \State $\mathcal{V} \gets \mathcal{V} \cup  \{s_k\}$ 
            \State Update $D$ by replacing all occurrences of $(s_i, s_j)$ with $s_k$ 
            \State Update frequencies of adjacent symbol pairs in $D$
        \EndWhile
        \State \Return $\mathcal{V}$ 
    \end{algorithmic}
\end{algorithm}

\section{Result for Translation Task}
\label{sec:appendix2}
To investigate how vocabulary size affects machine translation performance, we conduct experiments on three language pairs (German-English, French-English, and Chinese-English) . Each model variant is fine-tuned three times with different random seeds, and the average BLEU score is reported in Table~\ref{tab:trans_results}

\begin{table*}[ht]
    \centering
    \scriptsize  
    \setlength{\tabcolsep}{6pt}
    \renewcommand{\arraystretch}{0.9}
    \resizebox{\textwidth}{!}{
    \begin{tabular}{l|ccc|ccc|ccc}
        \toprule
        \textbf{Vocab} & \textbf{De-En} & \textbf{En-De} & \textbf{$R^2$} & \textbf{Fr-En} & \textbf{En-Fr} & \textbf{$R^2$} & \textbf{Zh-En} & \textbf{En-Zh} & \textbf{$R^2$} \\
        \textbf{size} & \textbf{BLEU} & \textbf{BLEU} &  & \textbf{BLEU} & \textbf{BLEU} &  & \textbf{BLEU} & \textbf{BLEU} & \\
        \midrule
        2,000   & 13.67 & 12.83 & 0.5755 & 15.01 & 15.67 & 0.5976 & 8.25  & 8.81  & 0.6668 \\
        5,000   & 19.31 & 17.99 & 0.6784 & 22.44 & 23.01 & 0.6940 & 12.91 & 12.92 & 0.6372 \\
        8,000   & 21.53 & 20.84 & 0.7266 & 25.46 & 27.08 & 0.7259 & 14.62 & 15.28 & 0.6153 \\
        10,000  & 23.33 & 22.86 & 0.7528 & 29.78 & 29.73 & 0.7477 & 16.15 & 16.74 & 0.6031 \\
        20,000  & 26.26 & 24.27 & 0.8136 & 33.71 & 35.40 & 0.8306 & 16.92 & 19.61 & 0.6678 \\
        30,000  & 27.22 & 25.11 & 0.8609 & 35.33 & 35.97 & 0.8909 & 18.92 & 21.80 & 0.7602 \\
        40,000  & 28.13 & 24.99 & 0.8968 & 36.48 & 36.95 & 0.9201 & 19.16 & 21.90 & 0.8196 \\
        50,000  & 28.88 & 25.78 & 0.9204 & 36.34 & 37.22 & 0.9397 & 18.78 & 22.34 & 0.8568 \\
        60,000  & 29.75 & 26.66 & 0.9382 & 36.57 & 37.37 & 0.9510 & 19.57 & 23.03 & 0.8846 \\
        70,000  & 29.80 & \textbf{26.89} & 0.9521 & \textbf{37.59} & 37.94 & 0.9642 & 19.91 & 22.95 & 0.9042 \\
        80,000  & \textbf{30.01} & 26.67 & 0.9609 & 36.89 & 38.06 & 0.9687 & 19.82 & 23.29 & 0.9144 \\
        100,000 & 29.99 & 26.52 & 0.9615 & 37.11 & \textbf{38.09} & 0.9622 & 20.51 & 23.94 & 0.9372 \\
        110,000 & 29.91 & 26.63 & 0.9648 & 37.10 & 38.15 & 0.9649 & 20.91 & 24.48 & 0.9578 \\
        120,000 & 29.86 & 26.58 & 0.9657 & 37.09 & 38.07 & 0.9680 & 20.92 & 24.40 & 0.9630 \\
        130,000 & 29.77 & 26.67 & 0.9625 & 37.29 & 38.03 & 0.9674 & \textbf{21.21} & \textbf{24.69} & 0.9596 \\
        140,000 & 29.69 & 26.62 & 0.9613 & 37.01 & 37.90 & 0.9636 & 21.00 & 24.52 & 0.9585 \\
        \bottomrule
    \end{tabular}}
    \caption{BLEU scores of models with different vocabulary sizes on the En-De, En-Fr, and En-Zh translation tasks. Each configuration is averaged over three random seeds.}
    \label{tab:trans_results}
    \vspace{-0.1in}
\end{table*}

\section{License and Terms of Use}
\label{sec:license}

We provide here the license information and terms of use for all datasets, models, and other artifacts used or created in this work.

\paragraph{Pre-training Datasets.}
\begin{itemize}
    \item \textbf{OpenWebText} and \textbf{BookCorpus} were used to pre-train BERT in the NLP domain. OpenWebText is a publicly available dataset intended to replicate the quality of OpenAI's WebText corpus and is distributed under an open research license.\footnote{\url{https://skylion007.github.io/OpenWebTextCorpus/}} BookCorpus was originally collected by \citet{zhu2015aligning} and is available for academic use only.
    \item \textbf{WMT16/17/18} datasets are used for multilingual pre-training and translation fine-tuning with mBART. These datasets are publicly released as part of the WMT shared tasks, licensed for research use.\footnote{\url{http://www.statmt.org/wmt16/}} 
    \item \textbf{ZINC20} is used for pre-training in the chemistry domain. ZINC is a free database of commercially-available compounds provided by the Irwin and Shoichet Laboratories at UCSF. It is available for academic research under a public domain dedication (CC0).\footnote{\url{https://zinc20.docking.org/}} 
    \item \textbf{DNA sequences} used for genomics tasks are derived from public genome datasets and follow the same data sources as DNABERT2 \cite{zhou2024dnabert}. These datasets are in the public domain and used solely for academic research.
\end{itemize}

\paragraph{Downstream Task Datasets.}
\begin{itemize}
    \item \textbf{GLUE Benchmark} datasets \cite{wang2018glue} are publicly released for research use and are commonly used under their respective licenses.
    \item \textbf{IWSLT14/15/17} datasets used for fine-tuning translation tasks are distributed for non-commercial research use as part of the IWSLT shared tasks.
    \item \textbf{MoleculeNet} datasets (e.g., BBBP, Tox21, Sider, ClinTox, HIV, BACE) are released under the MIT license and made publicly available by DeepChem.\footnote{\url{https://moleculenet.org/}}
    \item \textbf{GUE Dataset} used for genomics classification tasks is adopted following the usage in DNABERT2 \cite{zhou2024dnabert}, and is used for research purposes.
\end{itemize}

\paragraph{Code and Models.}
Our tokenizer construction scripts, Zipfian analysis tools, and vocabulary selection framework will be released under the MIT license. Any pre-trained models provided as part of this work will be licensed for academic research use only.

\section{Experimental Details}
\label{sec:exp_details}

\paragraph{Computational Resources.} 
All experiments were conducted using a combination of 8 NVIDIA 2080Ti GPUs and 4 NVIDIA A10 GPUs. In total, our experiments consumed approximately 4,900 GPU-hours on 2080Ti and 2,400 GPU-hours on A10 cards. These computations include all pre-training, fine-tuning, hyperparameter search, and validation runs across all domains.

\paragraph{Model Sizes.}
The number of parameters used in each experimental setting is summarized below:
\begin{itemize}
    \item \textbf{NLP (BERT models):} 84M to 124M parameters depending on vocabulary size.
    \item \textbf{NLP (mBART models):} 177M to 320M parameters depending on vocabulary size.
    \item \textbf{Genomics (BERT-based):} 80M to 93M parameters depending on vocabulary size.
    \item \textbf{Chemistry (BERT-based):} 72M to 90M parameters depending on vocabulary size.
\end{itemize}

\paragraph{Reproducibility.}
Each experiment was repeated using 3 different random seeds, and all reported results are averages over these runs.

\paragraph{Software and Libraries.}
We implemented all models using the HuggingFace Transformers library (v4.38) and PyTorch (v2.0). Data loading and pre-processing were done using the HuggingFace Datasets library. Evaluation metrics such as BLEU and ROC-AUC were computed using \texttt{nltk}, \texttt{scikit-learn}, and custom scripts, with standard configurations unless otherwise specified.

\end{document}